# A design oriented study for 3R Orthogonal Manipulators With Geometric Simplifications

Mazen ZEIN, Philippe WENGER and Damien CHABLAT
Institut de recherche en communications et cybernétique de Nantes UMR CNRS 6597
1, rue de la Noë, BP 92101, 44321 Nantes Cedex 03, France
Mazen.Zein@irccyn.ec-nantes.fr

*Abstract*— **This paper proposes a method to calculate the largest Regular Dextrous Workspace (RDW) of some types of three-revolute orthogonal manipulators that have at least one of their DH parameters equal to zero. Then a new performance index based on the RDW is introduced, the isocontours of this index are plotted in the parameter space of the interesting types of manipulators and finally an inspection of the domains of the parameter spaces is conducted in order to identify the better manipulator architectures.**

**The RDW is a part of the workspace whose shape is regular (cube, cylinder) and the performances (conditioning index) are bounded inside. The groups of 3R orthogonal manipulators studied have interesting kinematic properties such as, a well-connected workspace that is fully reachable with four inverse kinematic solutions and that does not contain any void.**

**This study is of high interest for the design of alternative manipulator geometries.**

I. INTRODUCTION

The workspace of general 3R manipulators has been widely studied in the past (see, for instance, [1-6]). The determination of the workspace boundaries, the size and shape of the workspace, the existence of holes and voids, the accessibility inside the workspace (i.e. the number of inverse kinematic solutions in the workspace), are some of the main features that have been explored.

Today, most industrial manipulators are of the PUMA type; they have a vertical revolute joint followed by two parallel joints and a spherical wrist. Another interesting category of serial manipulators exists, which have any two consecutive joint axes orthogonal. We call these manipulators *orthogonal manipulators*. Instances of orthogonal manipulators are the IRB 6400C launched by ABB-Robotics in 1998 and the DIESTRO manipulator built at McGill University.

Unlike PUMA type manipulators, orthogonal manipulators may have many different kinematic properties according to their links and joint offsets lengths. Orthogonal 3R manipulators may be binary (only two inverse kinematic solutions) or quaternary (four inverse kinematic solutions), they may have voids or no voids in their workspace, they may be cuspidal or noncuspidal.

In [7], the authors attempted the classification of 3R orthogonal manipulators with no offset on their last joint. The work of [7] was completed in [8] where the authors established a classification of two families of 3R orthogonal manipulators, according to the number of cusps and node points. The parameter space of each family was divided into a number of cells where the manipulators have the same number of cusps and nodes in their workspace.

More recently, in [9] about ten remaining families of 3R orthogonal manipulators with at least one of their DH parameters equal to zero have been classified into different types with similar singularities topologies and similar kinematic properties. The resulting classification has shown that five manipulators types have a well-connected workspace, which was shown in the past to be specific to Puma-type manipulators. These types are interesting candidates for the design of alternative manipulator geometries. The classification of [9] relies only on the topology of the workspace. Other interesting kinematic features, such as the compactness of the workspace or the global conditioning index, were not analyzed.

In this paper, we will continue the work of [9] by taking into account these last two features, we will consider a new feature the regular dextrous workspace (RDW), which combines the last two features cited. The notion of RDW was first introduced in [10]. The RDW is a part of the workspace whose shape is regular (cube, cylinder) and the performances (conditioning index) are bounded inside.

Next section of this paper presents the families of manipulators under study and recalls some preliminary results. A method to calculate the RDW is presented in section 3. In section 4, a new performance index is introduced, it is the ratio between the length of the RDW edge and the maximal reach, the isocontours of this new performance index are computed in the parameter space of each type of the 3R manipulators studied, and then an inspection of the parameter spaces is conducted in order to find the domains where the manipulators have the best ratio and to identify the better types of manipulators regarding this performance index. Section 5 concludes this paper.

## II. PRELIMINARIES

### A. Orthogonal manipulators

The manipulators studied, referred to as *orthogonal manipulators*, are positioning manipulators with three revolute joints in which the two pairs of adjacent joint axes are orthogonal. The length parameters are $d_2$, $d_3$, $r_2$, $r_3$ and $d_4 > 0$. The angle parameters $\alpha_2$ and $\alpha_3$ are set to $-90°$ and $90°$, respectively. The three joint variables are referred to as $\theta_1$, $\theta_2$ and $\theta_3$, respectively. They will be assumed unlimited in this study. The position of the end-tip is defined by the three Cartesian coordinates $x$, $y$ and $z$ of the operation point P with respect to a reference frame (O, **X**, **Y**, **Z**) attached to the manipulator base. Figure 1 shows the architecture of the manipulators under study in the home configuration defined by $\theta_1 = \theta_2 = \theta_3 = 0$.

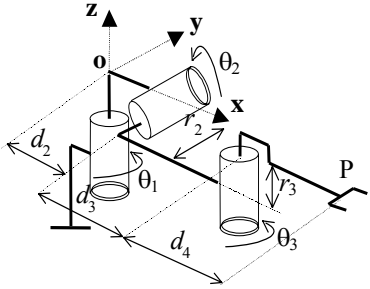

Figure 1. A 3R orthogonal manipulator.

### B. Orthogonal manipulators with geometric simplifications

In the classification provided in [9], all the possible combinations of manipulators with at least one DH parameter equal to zero were treated. The different combinations are depicted by the tree shown in Fig. 2, which yields ten families. Parameter $d_4$ cannot be equal to zero because the resulting manipulator would be always singular.

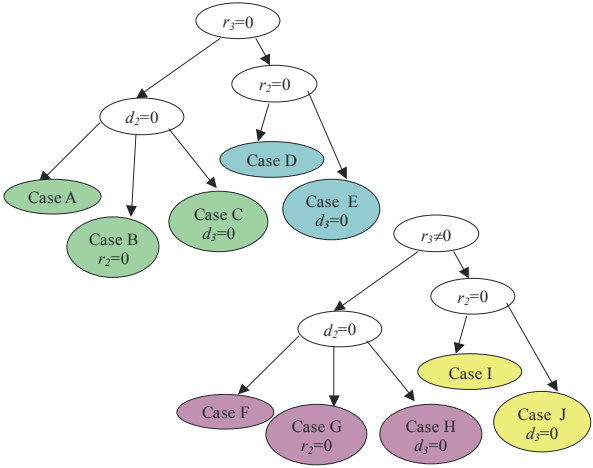

Figure 2. The ten families of 3R orthogonal manipulators with at least one parameter equal to zero

### C. Classification of 3R orthogonal manipulators with geometric simplifications

In [9], an exhaustive classification and enumeration of all types of workspace topology was conducted for the ten families of orthogonal manipulators that have at least one geometric parameter equal to zero. Twenty-two different types of manipulators were identified, which have similar global kinematic properties.

For all manipulators of one given type, the following global kinematic properties are the same: (i) number of nodes (ii) number of voids (iii) number of 2-solution and 4-solutions regions (iv) t-connectivity and well-connectivity of the workspace. Five types of manipulators were identified to have interesting kinematic properties, namely types B1, C, E, G and H. Each type has the same parameters properties defined in Figure 2, and beside that the manipulators of type B1 have $d_3 > d_4$. The manipulators of these five types have a well-connected workspace (workspace is fully reachable with four inverse kinematic solutions and fully t-connected). Figure 3 shows a workspace section of a manipulator pertaining to each of the five types.

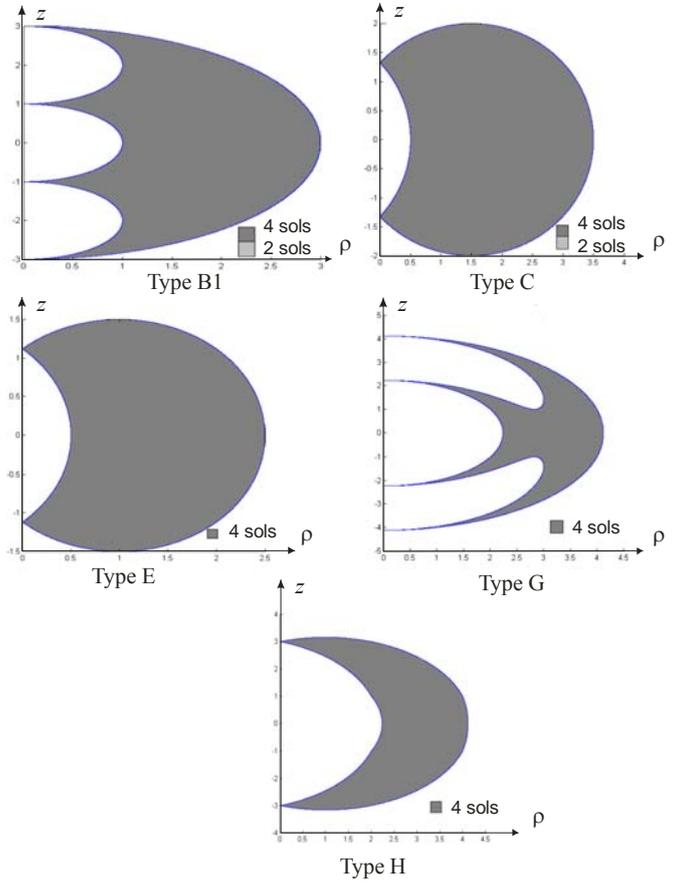

Figure 3. The five manipulator types with well-connected workspace.

### D. Performance indices

Several criteria have been used in the past to compare robotic manipulators, such as the workspace [1-6] or the conditioning index [11-13].

The conditioning index gives a measure of the accuracy of (*i*) the Cartesian velocity of the end effector that is produced by the joint rates calculated from Jacobian inversion and (*ii*) the static load (force and moment) acting on the end effector upon measurements by torque cells at the joint axes.

The conditioning index *k* is defined as the ratio between the maximal and minimal eigenvalues of the Jacobian matrix J. For representation reasons, in our paper the conditioning index will be $k^{-1}$ (the inverse of *k*):

$$k^{-1} = \sigma_{min} / \sigma_{max} \quad (1)$$

In our study, to compare the performances of the 3R orthogonal manipulators, we will use an original and novel performance index $\eta$ combining the two performance indices cited before (workspace and conditioning index), this index will be based on the notion of the regular dextrous workspace RDW.

The RDW is a part of the workspace whose shape is regular and the performances (conditioning index $k^{-1}$ in our study) are bounded inside.

The performance index $\eta$ will be the ratio between the edge's length of the RDW and the maximal reach of the manipulator.

### III. APPROACH TO CALCULATE THE RDW

In a context of design and/or trajectory planning, an important problem is to find singularity-free zones in the workspace of a manipulator.

In this section, we introduce a new procedure to determine the RDW of 3R orthogonal manipulators. A numerical example is provided in order to illustrate the effectiveness of the procedure.

Because the workspace is symmetric about the first joint axis, a section of the plane ($\rho = \sqrt{x^2+y^2}$, *z*) is sufficient to study the 3D workspace. The RDW is thus chosen as a planar regular shape. We choose a square but other choices would be possible, such as a rectangle or a disc.

Our approach to calculate the RDW will be divided into two parts, the first one consists in determining the regular workspace section, which is the largest singularity free square in a half cross-section of the workspace, the second part consists in determining the RDW section which is a square situated inside the regular workspace section (computed in the first part), and where all the points inside have a conditioning index greater than a fixed value.

#### A. Regular workspace

For a better comprehension of the procedure aiming at calculating the regular workspace, we illustrate it directly on a type C 3R orthogonal manipulator having the following non-zero parameters: $d_4=1.5$ and $r_2=1$. Figure 4 shows a workspace section of this manipulator.

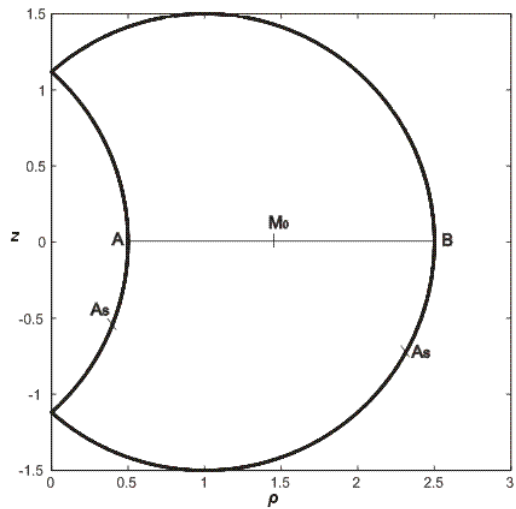

Figure 4. Workspace section of a type C manipulator

Determining the regular workspace of this manipulator consists in finding the largest singularity free square inside the workspace section, which is described by the following approach:

***Step 1:*** We compute the coordinates of the middle point $M_0(\rho_0,0)$ (Fig.4) of the linear segment AB on the z=0 axis, where A and B are the first singular points met from $M_0$. This point will constitute the initial center of the singularity free square. We have chosen $M_0$ to be on the z=0 axis because of the symmetry of the workspace about this axis (this symmetry is due to the fact the manipulators studied are orthogonal).

***Step 2:*** We calculate the largest singularity free square centered at $M_0(\rho_0,0)$.

To do this, we calculate the infinity norm distance *d*, also known as the Chebyshev distance, between the center point $M_0(\rho_0,0)$ and each of singular points $A_s(\rho_s,z_s)$ (where $A_s$ represents all the singular points of the workspace section Fig.4).

$$d = \max\left(|\rho_0 - \rho_s|, |z_0 - z_s|\right) \quad (2)$$

and we keep the minimal distance $d_{min}$ found over all, because we are searching for the distance between the closest singularity configuration $A_s$ from the center point $M_0$. The edge length of the largest singularity free square will be twice $d_{min}$.

***Step 3:*** The choice of the initial center point $M_0(\rho_0,0)$ does not lead to an optimized solution, in other words varying lightly the center point position may lead to a largest singularity-free square. Thus, the position of the initial point must be optimized, which we have done using a Hooke and Jeeves optimization scheme [14], to obtain finally a point M center of the largest singularity free square.

The length of the singularity free square edge *a* will be:

$$a = 2 \times d_{min} \quad (3)$$

Figure 5 shows the largest singularity free square (displayed in hatched) inside the workspace section of the same manipulator used in Figure 4.

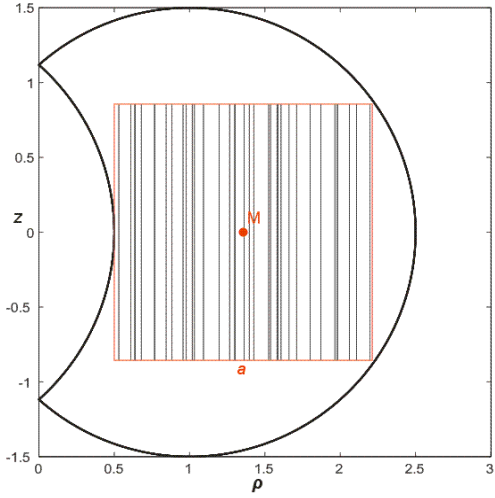

Figure 5. Largest singularity free square inside the workspace section of a type C 3R orthogonal manipulator

To obtain the whole regular workspace it suffices to rotate the square of 360° about the z axis.

*B. Regular dextrous workspace*

To determine the RDW, we have first to fix a minimal value $k^{-1}_{min}$ of the conditioning index.

***Step 4:*** We perform a regular scan of points inside the singularity free square computed in *step3*.

We calculate the conditioning index $k^{-1}_{scan}$ at each of the points scanned. To do this, we calculate the pre-images through the inverse kinematics of each point scanned, and then we calculate the conditioning index at each inverse kinematic solution.

For the manipulators of type B1, C, E, G and H the conditioning index is the same for all the inverse kinematic solutions, because these manipulators do not have singularities inside their workspace, so their workspace are t-connected for all the aspects (see [9]). The workspace of a manipulator is said to be *t-connected* if any continuous trajectory is feasible throughout [15].

So, if the manipulator studied is of the type B1, C, E, G or H, the conditioning index $k^{-1}_{scan}$ of a point scanned is the conditioning index of one of its inverse kinematic solution, else the conditioning index $k^{-1}_{scan}$ can be chosen to be the maximal conditioning index of the inverse kinematic solutions.

***Step 5:*** We take a square centered at the point M (computed in *step 3*) having an edge length $a_\varepsilon$, where $a_\varepsilon$ is equal to the scanning step used in step 4. We test if the points scanned in *step 4* and which pertain to this square have all a conditioning index $k^{-1}_{scan} > k^{-1}_{min}$, if yes we increase the value of $a_\varepsilon$ and we repeat the same test.

We still increase the value of $a_\varepsilon$ until we obtain a square containing a scanned point having $k^{-1}_{scan} < k^{-1}_{min}$.

The choice of the center point M does not lead to an optimized solution, in other words varying lightly the center point position may lead to a largest section of the RDW. Thus, the position of the initial point must be optimized, which we have done using a Hooke and Jeeves optimization scheme [14], to obtain finally a point $M_{RDW}$ center of the largest square section of the RDW having an edge length $a_{RDW}$.

Figure 6 shows in blue, for $k^{-1}_{min}=0.25$, the largest square section of the RDW inside the workspace section of the same manipulator used in Figure 4 (the largest singularity free square is also represented in hachured).

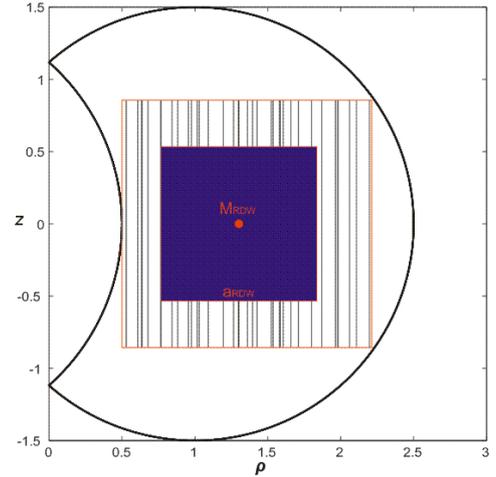

Figure 6. Largest RDW square section for $k^{-1}_{min}=0.25$ inside the workspace section of a type C 3R orthogonal manipulator

To obtain the whole regular dextrous workspace it suffices to rotate the square of 360° about the z axis.

This approach is of high interest for manipulator design and for trajectory planning.

IV. COMPARISON OF 3R ORTHOGONAL MANIPULATORS

In [9], ten families of 3R orthogonal manipulators were classified. As a consequence twenty-two different types of manipulators were found, which have similar global kinematic properties. Finally, just five of these types of manipulators were identified to have very interesting kinematic properties (such as well-connected workspace).

In this section, these five types of manipulators will be compared according to the novel performance index $\eta$ introduced in section II-D.

The performance index $\eta$ is the ratio between the edge's length of the RDW section and the maximal reach of the manipulator.

$$\eta = \frac{a_{RDW}}{\rho_{max}} \quad (4)$$

This performance index combines *(i)* the compactness of the workspace because it considers a characteristic length of a singularity free regular surface inside the workspace also it considers the maximal reach; *(ii)* the conditioning of the workspace because the conditioning index of the configurations inside the RDW are higher than a fixed minimum value $k^{-1}_{min}$. So, a manipulator having a high performance index $\eta$ should have a compact and well-conditioned workspace.

So to compare these five types of manipulators, we will proceed first by scanning the parameter space of each of the five types, and by calculating the performance index $\eta$ at each point scanned (each point in the parameter space represents a manipulator). Then after calculating the performance index $\eta$ of each scanned point in the parameter space, we plot the isocontours of $\eta$ in the parameter space of each of the five types of 3R orthogonal manipulators.

In our study we have chosen $k^{-1}_{min}$ to be 0.25, we show below the application of this procedure on each of the five types.

*A. Type B1*

The manipulators of this type have $d_3 \neq 0$, $d_4 \neq 0$ and $d_3 > d_4$. Their parameter space is two-dimensional ($d_3$, $d_4$). Figure 7 shows the parameter space of type B manipulators, the transition curve (d) between types B1 and B2 manipulators, and the isocontours of the performance index $\eta$ for $k^{-1}_{min} = 0.25$. We can see in figure 7 that the highest value of $\eta$ is 0.5. We can see also that the manipulators having a high performance index 0.5 are such that $d_3 \approx 1.5 d_4$.

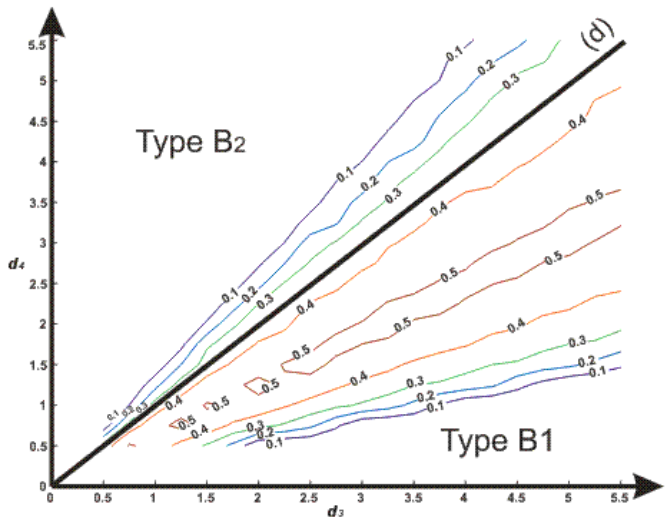

Figure 7. Parameter space and isocontours of the performance index $\eta$ of the type B manipulators.

*B. Type C*

The manipulators of this type have $r_2 \neq 0$ and $d_4 \neq 0$. Their parameter space is two-dimensional ($r_2$, $d_4$). Figure 8 shows the parameter space of type C manipulators, and the isocontours of the performance index $\eta$ for $k^{-1}_{min} = 0.25$. We can see in figure 8, that the highest value of $\eta$ is 0.55, which is better than B1's manipulators highest value, we can see also that the manipulators having a high performance index (0.5 and 0.55) are situated near the bisector of the parameter space, which means that type-C manipulators with maximal $\eta$ are such that $r_2 \approx d_4$.

By comparing, $B_1$'s and C's isocontours in the parameter space, we can remark that the area occupied by the isovalues 0.5 and 0.55 in C's parameter space is largest than the one of the isovalues 0.5 in $B_1$'s parameter space.

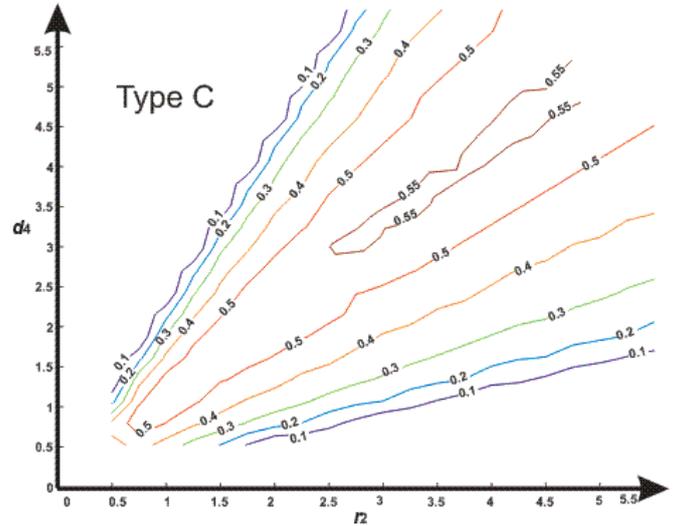

Figure 8. Parameter space and isocontours of the performance index $\eta$ of the type C manipulators.

*C. Type E*

The manipulators of this type have $d_2 \neq 0$ and $d_4 \neq 0$. Their parameter space is two-dimensional ($d_2$, $d_4$). Figure 9 shows the parameter space of type E manipulators, and the isocontours of the performance index $\eta$ for $k^{-1}_{min} = 0.25$. We can see in figure 9, that the highest value of $\eta$ is 0.4, which is lower than B1's and C's manipulators highest value. So we can say the type B1 and C manipulators are better than type E ones.

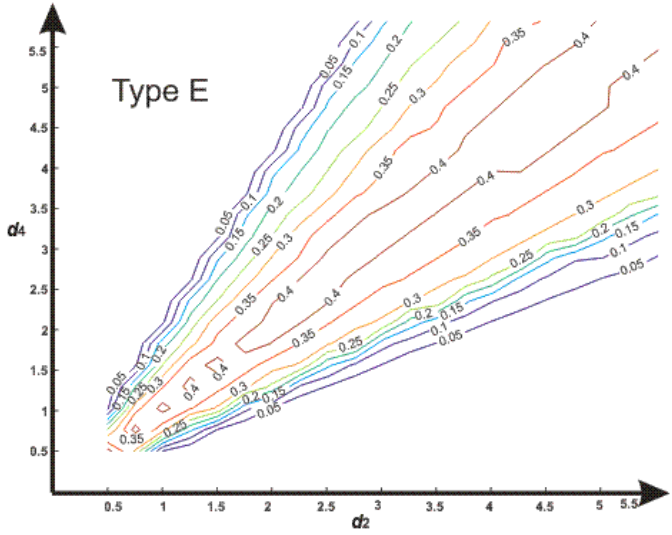

Figure 9. Parameter space and isocontours of the performance index *η* of the type E manipulators.

### D. Type G

The manipulators of this type have $d_3 \neq 0$, $d_4 \neq 0$ and $r_3 \neq 0$. We normalize the three non-zero parameters by $r_3$ in order to reduce the dimension of the problem. To do this we fix the value of $r_3$ to 1. So, the parameter space is two-dimensional ($d_3$, $d_4$) for $r_3 = 1$. Figure 10 shows the parameter space of type G manipulators, and the isocontours of the performance index *η* for $k^{-1}_{min} = 0.25$. We can see in figure 10 that the highest value of *η* is 0.5. We can see also that the area occupied by the isovalues 0.5 is smaller than the one of the isovalues 0.5 in type C's parameter space.

We notice that type G manipulators are close to type B1 ones regarding their performance index *η*. Also, the type G manipulators having a high performance index 0.5 are such that $d_3 \approx 1.5 d_4$.

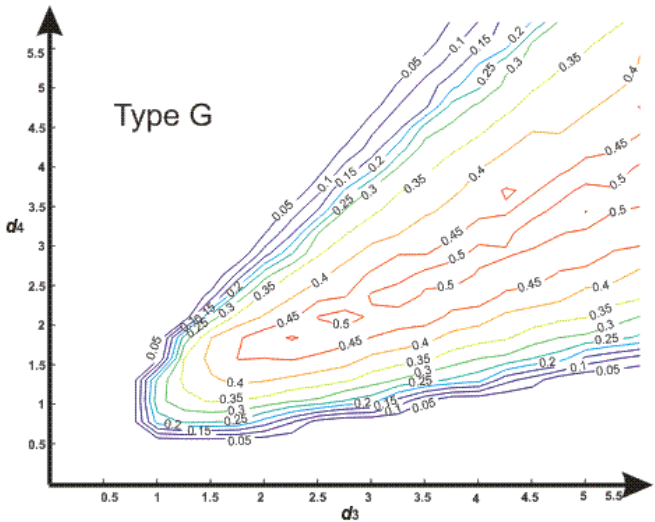

Figure 10. Parameter space and isocontours of the performance index *η* of the type G manipulators.

### E. type H

The manipulators of this type have $r_2 \neq 0$, $d_4 \neq 0$ and $r_3 \neq 0$. We normalize the three non-zero parameters by $r_3$ in order to reduce the dimension of the problem. To do this we fix the value of $r_3$ to 1. So, the parameter space is two-dimensional ($r_2$, $d_4$) for $r_3 = 1$. Figure 11 shows us the parameter space of type H manipulators, and the isocontours of the performance index *η* for $k^{-1}_{min} = 0.25$. We can see in figure 11 that the highest value of *η* is 0.55. We can see also that the area occupied by the isovalues 0.5 and 0.55 is too close to the one of the isovalues 0.5 and 0.55 in type C's parameter space. Also, the type H manipulators having a high performance index (0.5 and 0.55) are such that $r_2 \approx d_4$.

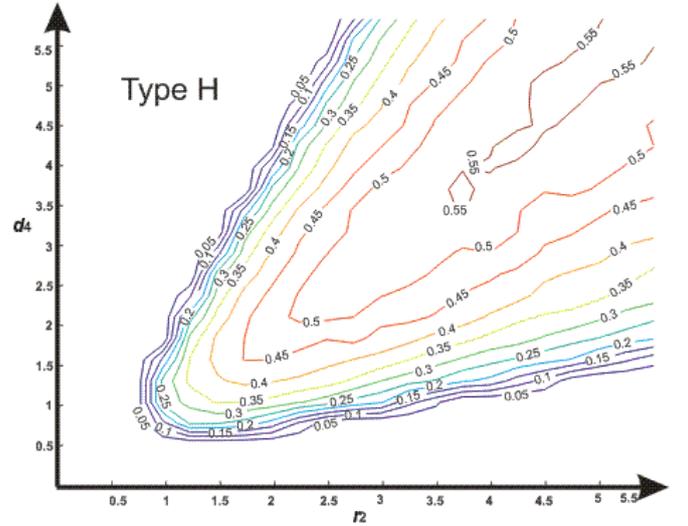

Figure 11. Parameter space and isocontours of the performance index *η* of the type H manipulators.

### F. Conclusion

Above we have seen that the manipulators of type C and H can have the highest performance index *η* which is 0.55 over all the other types, we have seen also that the manipulators of types C and H having a *η* higher than 0.5 and 0.55 cover a large area of the parameter space of C and H, which is not the case for the other manipulator types.

By comparing the five types $B_1$, C, E, G and H of 3R orthogonal manipulators regarding the performance index *η*, we conclude that type C and H manipulators are the best ones. We can say as a consequence that types C and H manipulators are the best ones to combine both a compact and a well conditioned workspace.

Plotting the isocontours in the parameter spaces constitutes an important tool that helps choosing manipulators having both a compact and a well conditioned workspace.

Figure 12 shows a representative manipulator morphology of each type and the workspace and the RDW of a manipulator having the highest performance index of each type. These manipulators were chosen using the parameter space isocontours of each type.

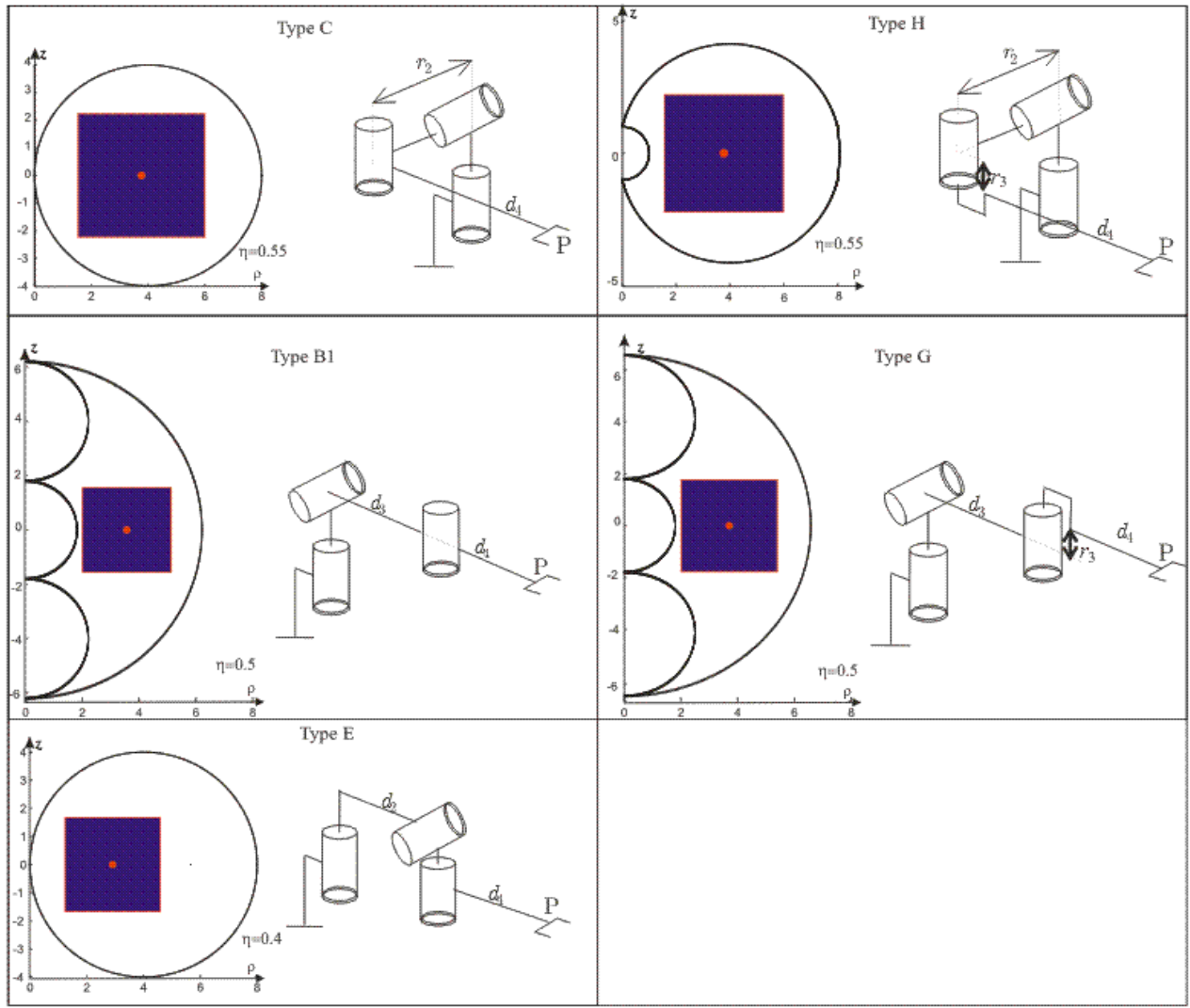

Figure 12. Workspaces and RDW of the manipulators having the following parameters: Type B1 ($d_3$=4 and $d_4$=2.2); Type C ($d_4$=4 and $r_2$=4); Type E ($d_2$=4 and $d_4$=4); Type G ($d_3$=4, $d_4$=2.5 and $r_3$=1); Type H ($r_2$=4, $d_4$=4 and $r_3$=1).

## V. Conclusion

In this paper, a procedure for the determination of the regular workspace and the Regular Dextrous Workspace (RDW) has been provided.

A new performance index has been introduced; this performance index combines both the compactness and the conditioning of the workspace. It is the ratio between the section edge length of the RDW and the maximal reach of the manipulator.

The isocontours of the new performance index $\eta$ were plotted in the parameter space of each of the five types.

Finally, by interpreting the isocontours obtained, we have concluded that types C and H are the better ones regarding the performance index $\eta$. The manipulators of these types combine both a compact and a well conditioned workspace.